%% file: root.tex
\documentclass[letterpaper, 10 pt, conference]{ieeeconf}  

\IEEEoverridecommandlockouts                              

\usepackage{graphics} 
\usepackage{epsfig} 
\usepackage{mathptmx} 
\usepackage{times} 
\usepackage{amsmath} 
\usepackage{amssymb}  

\usepackage{hyperref}
\usepackage{url}
\usepackage{tabularx}
\usepackage{multirow}
\usepackage{booktabs}
\usepackage{pifont}
\usepackage{algorithm}
\usepackage{algorithmic}
\usepackage{makecell}
\usepackage{xcolor}
\usepackage{caption}
\usepackage{siunitx}
\usepackage{xspace}

\renewcommand{\baselinestretch}{0.98}

\newcommand{\OmniRetarget}{\textsc{OmniRetarget}\xspace}

\title{\LARGE \bf
OmniRetarget: Interaction-Preserving Data Generation for Humanoid Whole-Body Loco-Manipulation and Scene Interaction
}

\author{Lujie Yang*$^{1, 2}$, Xiaoyu Huang*$^{1, 3}$, Zhen Wu*$^{1}$, \\ Angjoo Kanazawa$^{\dagger 1, 3}$, Pieter Abbeel$^{\dagger 1, 3}$, Carmelo Sferrazza$^{\dagger 1}$, C. Karen Liu$^{\dagger 1, 4}$, Rocky Duan$^{\dagger 1}$, Guanya Shi$^{\dagger 1, 5}$ \\
$^{1}$Amazon FAR (Frontier AI \& Robotics), $^{2}$MIT, $^{3}$UC Berkeley, $^{4}$Stanford University, $^{5}$CMU %
\thanks{* Equal contribution, work done while interning at Amazon FAR. $\dagger$ Amazon FAR team co-lead.}
}

\newcommand{\yes}{\textcolor{green}{\ding{51}}}
\newcommand{\no}{\textcolor{red}{\ding{55}}}

\DeclareMathOperator*{\argmin}{arg\,min}

\AddToHook{cmd/@maketitle/after}{\ADDINITIALFIGURE}
\newcommand{\ADDINITIALFIGURE}{%
  \begin{minipage}{\textwidth}
    \expandafter\def\csname @captype\endcsname{figure}%
    \centering
    \includegraphics[width=\textwidth]{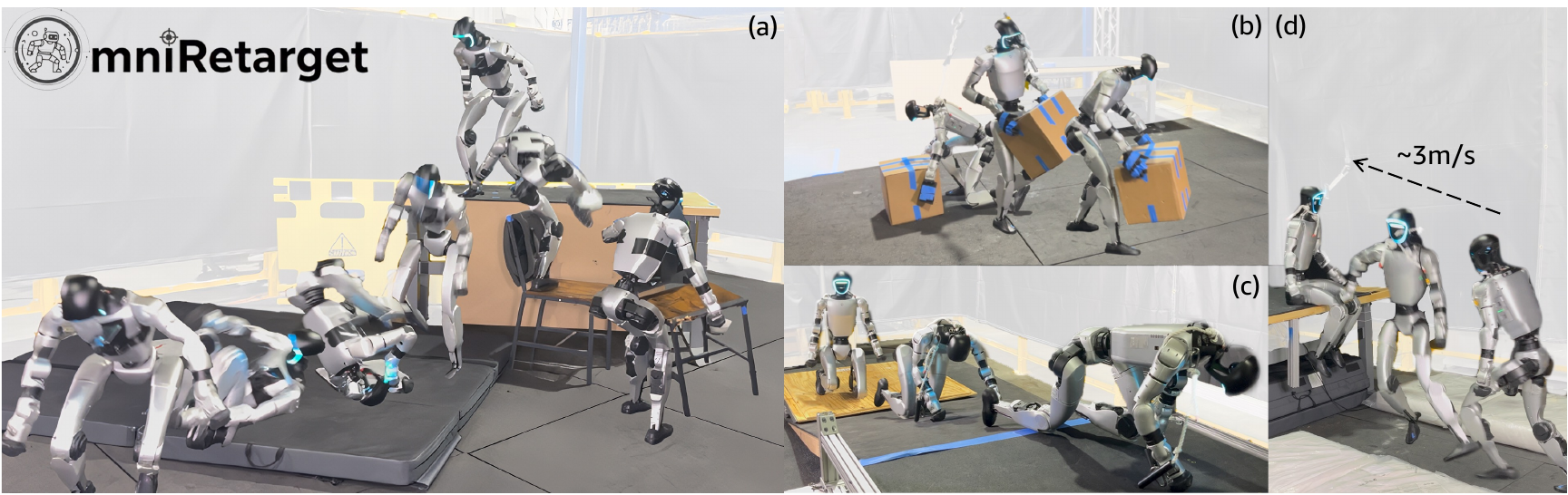}%
    \setcounter{figure}{0}
    \captionof{figure}{
    \OmniRetarget enables reinforcement learning policies to learn complex, long-horizon loco-manipulation skills in challenging environments that transfer zero-shot from simulation to a Unitree G1 humanoid. Thanks to the high-quality interaction-preserving motion retargeting, these policies are trained and deployed in a \emph{minimal and unified} way: it involves only 5 rewards, 4 robot domain randomization terms, and a purely proprioceptive observation space, shared by all tasks. Demonstrated behaviors include \textbf{(a)} 30-second parkour course involving chair moving, stepping \& vault, and jump \& roll, \textbf{(b)} object transportation, \textbf{(c)} crawling on a slope, and \textbf{(d)} fast platform climbing and sitting. 
    \label{fig:flagship_demo}}
  \end{minipage}}
  
\begin{document}

\maketitle
\thispagestyle{empty}
\pagestyle{empty}

\input{sections/00_abstract}
\input{sections/01_introduction}

\input{sections/02_related_works}
\input{sections/03_kinematic_retarget}

\input{sections/04_training}

\input{sections/05_experiments}
\input{sections/06_conclusion}

\begin{small}
\bibliography{IEEEabrv, root}
\bibliographystyle{IEEEtran}
\end{small}

\input{sections/07_appendix}
\end{document}

%% file: sections/00_abstract.tex
\begin{abstract}
A dominant paradigm for teaching humanoid robots complex skills is to retarget human motions as kinematic references to train reinforcement learning (RL) policies. However, existing retargeting pipelines often struggle with the significant embodiment gap between humans and robots, producing physically implausible artifacts like foot-skating and penetration. More importantly, common retargeting methods neglect the rich human-object and human-environment interactions essential for expressive locomotion and loco-manipulation.
To address this, we introduce \OmniRetarget, an interaction-preserving data generation engine based on an interaction mesh that explicitly models and preserves the crucial spatial and contact relationships between an agent, the terrain, and manipulated objects. By minimizing the Laplacian deformation between the human and robot meshes while enforcing kinematic constraints, \OmniRetarget generates kinematically feasible trajectories. Moreover, preserving task-relevant interactions enables efficient data augmentation, from a single demonstration to different robot embodiments, terrains, and object configurations. 
We comprehensively evaluate \OmniRetarget by retargeting motions from OMOMO \cite{li2023object}, LAFAN1 \cite{harvey2020robust}, and our in-house MoCap datasets, generating over 8-hour trajectories that achieve better kinematic constraint satisfaction and contact preservation than widely used baselines.
Such high-quality data enables proprioceptive RL policies to successfully execute long-horizon (up to 30 seconds) parkour and loco-manipulation skills on a Unitree G1 humanoid, trained with only 5 reward terms and simple domain randomization shared by all tasks, without any learning curriculum. 
All code, retargeted datasets, and trained policies will be publicly released. Result videos can be found at \href{https://omniretarget.github.io}{https://omniretarget.github.io}
\end{abstract}

%% file: sections/01_introduction.tex
\section{Introduction}
The quest to enable humanoid robots to perform complex whole-body scene- and object-interaction tasks has long been constrained by a fundamental data bottleneck. While deep reinforcement learning (RL) has shown remarkable success in robot control, efficient exploration is highly sensitive to reward engineering~\cite{lee2020learning}.
This challenge is further amplified on humanoids, whose high-dimensional action spaces and complex dynamics make learning natural, expressive behaviors from scratch both difficult and inefficient.


To address these challenges, imitating human motions offers a powerful alternative for learning whole-body control, especially for complex scene interactions. Human demonstrations capture dynamic coordination, such as lifting objects while walking on uneven terrain, and have been used effectively in animation~\cite{peng2018deepmimic, xu2025parc, wu2024human}. A critical challenge arises in robotics: unlike virtual characters, physical humanoids only approximate human morphology, with significant differences in shape, proportion and degrees of freedom. This embodiment gap means that simply adapting human motions is insufficient; it is essential to also adapt their scene interactions to the robot's specific form to generate usable references.

To this end, researchers have pursued two main strategies. The first one is teleoperation~\cite{fu2024humanplus, he2024omnih2o, ze2025twist}, where only a human operator's motions are retargeted to control the robot online. This approach leverages the human operator for real-time adaptation, which sidesteps the need for automatic interaction retargeting. However, despite the advantage of online feedback, the method remains labor-intensive and does not scale well for large-scale data generation. The second and more scalable strategy is offline interaction retargeting, which holistically adapts both the human's motion and their scene interactions to the robot's specific embodiment.

However, most existing retargeting methods~\cite{Luo2023PerpetualHC, ze2025twist, videomimic} fall short in this regard. They predominantly rely on unconstrained or softly-penalized optimization, resulting in implausible motions with artifacts such as foot skating and penetration. More importantly, they do not explicitly consider interaction preservation---i.e., maintaining spatial and contact relationship---in the retargeting formulation, relying instead on simple keypoint matching. Consequently, the resulting references are of lower quality, which in turn complicates the downstream RL policy training~\cite{zhang2025hub, he2024omnih2o, he2025asap}.

\begin{figure*}
\centering
\includegraphics[width=\textwidth]{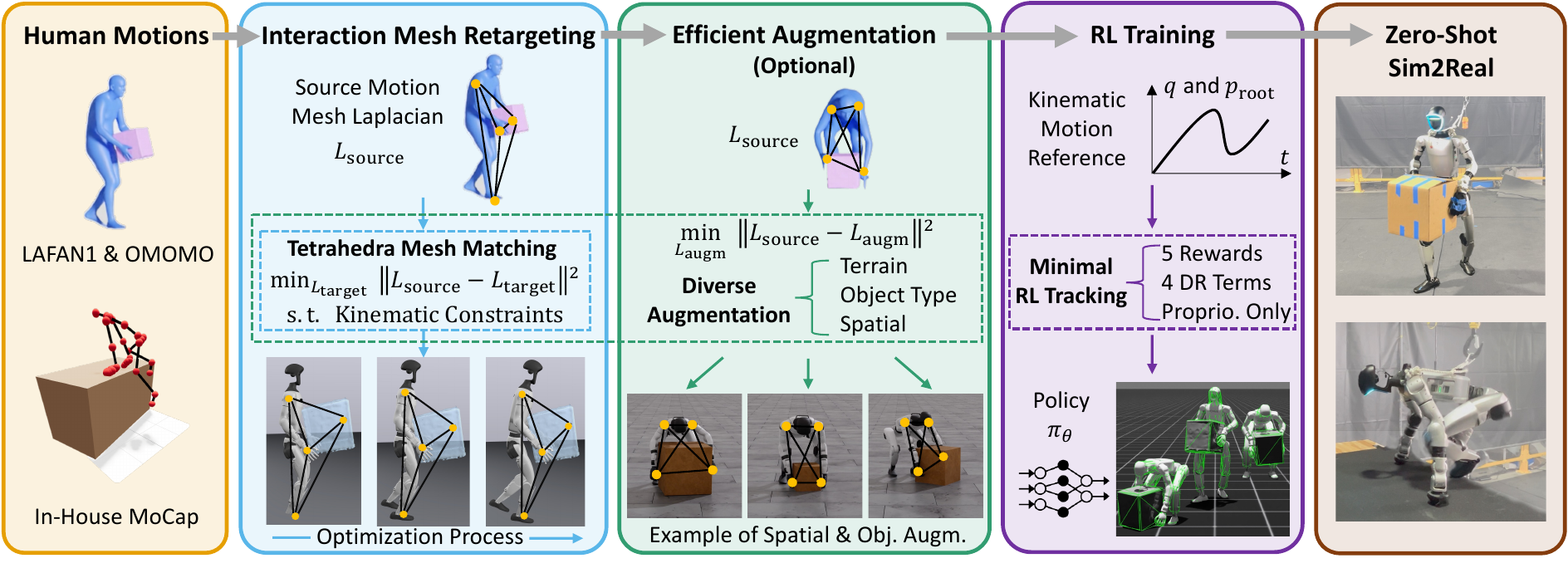}
	\caption{\textbf{\OmniRetarget overview.} Human demonstrations are retargeted to the robot via interaction-mesh–based constrained optimization. Each spatial and shape augmentation is solved as a new optimization, producing diverse trajectories that serve as references for RL training with minimal reward design and domain randomization, enabling zero-shot transfer to real-world humanoids.} 
	\label{fig:banner}
\end{figure*}

In this work, we introduce \OmniRetarget, an open-source data generation engine that transforms human demonstrations into diverse, high-quality kinematic references for humanoid whole-body control. By modeling spatial and contact relationships between robots, objects, and terrains via an interaction mesh \cite{Ho2010Spatial}, \OmniRetarget preserves essential interactions and generates kinematically feasible variations. While existing methods require separate demonstrations for each variation---making data collection costly and limiting coverage---\OmniRetarget addresses this bottleneck directly. Inspired by data augmentation frameworks for contact-rich manipulation~\cite{yang2025physics}, our framework automatically augments a single demonstration into a large number of training examples across object configurations, shapes, robot embodiments, and environments.

Our pipeline employs constrained optimization to enforce physical feasibility, including collision avoidance, joint limits, and foot contact stability, while minimizing interaction mesh deformation. The resulting motions are interaction-preserving and exhibit only minimal kinematic artifacts, providing dense learning signals that accelerate RL with minimal reward engineering. On a diverse suite of whole-body interaction tasks such as box lifting, platform climbing, and slope crawling, policies trained on \OmniRetarget datasets outperform those from prior retargeting methods in both motion quality and robustness, with successful zero-shot sim-to-real transfer onto a physical humanoid robot.

Our contributions are fourfold:
\begin{enumerate}
    \item The first interaction-preserving humanoid retargeting framework that handles rich robot-object-terrain interactions while enforcing hard physical constraints.
    \item A systematic data augmentation pipeline that transforms a single human demonstration into a diverse, large-scale set of high-quality kinematic trajectories on various robot embodiments.
    \item A large-scale, open-source dataset of retargeted, kinematically-feasible loco-manipulation trajectories.
    \item Successful zero-shot sim-to-real transfer of proprioceptive RL policies on a physical humanoid, demonstrating a diverse set of scene-interaction tasks, including a long, agile sequence of object carrying, platform climbing, jumping, rolling and wall-flipping.
\end{enumerate}

%% file: sections/02_related_works.tex
\section{Related Works}
\subsection{Motion Retargeting}\label{sec:motion_retargeting}
In computer graphics, transferring motions across characters has been extensively explored. Researchers have employed optimization-based methods to retarget human motions to avatars by preserving distances and orientations between keypoints \cite{cheynel2023sparse}, minimizing deformation energy \cite{Ho2010Spatial, kim2016retargeting}, or scaling the motions to satisfy hard constraints \cite{gleicher1998retargetting}. Others leverage data-driven methods that map diverse skeletons to a canonical representation \cite{aberman2020skeleton}, solve inverse kinematics with neural networks \cite{villegas2018neural}, or use reinforcement learning to preserve an interaction graph \cite{zhang2023simulation}.

Retargeting motions to humanoid robots introduces challenges beyond character animation, particularly the need to enforce physical constraints. For example, PHC \cite{Luo2023PerpetualHC}, a graphics method adopted in robotics~\cite{he2025asap, he2024omnih2o}, uses keypoint matching with unconstrained optimization, often leading to penetration, foot skating, and lack of object or scene awareness.
Similarly, GMR~\cite{ze2025twist} extends keypoint matching to orientations but suffer the same issues. 
VideoMimic~\cite{videomimic} improves realism with soft contact and collision penalties but offers no guarantees and requires careful tuning.

The closest method to ours is Interaction Mesh based Motion Adaptation (IMMA) \cite{Nakaoka2012Interaction}, which also leverages an interaction mesh \cite{Ho2010Spatial} to preserve the spatial relationship between body parts. However, it is not open-sourced and ignores kinematic limits and interactions with the environment or manipulated objects. 
In contrast, \OmniRetarget unifies all hard constraints, including foot sticking, non-penetration, and joint and velocity limits, while explicitly preserving environment and object interactions.

\subsection{Learning-Based Humanoid Whole-Body Control}

Recent learning-based whole-body control has enabled humanoids to traverse dynamic scenes and manipulate objects~\cite{dao2024sim, long2024learning, he2025attention, he2025learning, kuang2025skillblender, zhang2025unleashing, xue2025unified, zhang2406wococo, zhang2025falcon}. These methods typically train with hand-crafted rewards or task interfaces (e.g., velocity tracking, contact schedules, end-effector targets) but depend on extensive reward engineering and mostly fail to yield natural, human-level motions.

Motion imitation offers a promising alternative. In graphics, DeepMimic~\cite{peng2018deepmimic} shows that using human references yields natural, human-like behaviors with agile, dynamic motions. 
However, applying this approach to humanoid robots remains difficult due to the lack of reliable open-source kinematic retargeting pipelines.
With suboptimal reference motions, practitioners are forced to either manually clean the data~\cite{zhang2025hub} or re-introduce extensive reward engineering, such as ad-hoc regularizers for contact, slipping, and air time, to compensate for artifacts~\cite{ze2025twist, he2025asap, li2025reinforcement}. In contrast, trackers with minimal reward formulation like BeyondMimic~\cite{liao2025beyondmimic} achieve state-of-the-art results on hardware with high-fidelity references~\cite{unitree_lafan1_retargeting_dataset}, but those are scarce and robot-only, without interactions.

Beyond single-character motion, human–scene interaction data has proven effective for terrain traversal and loco-manipulation in character animation~\cite{xu2025parc, wu2024human, xu2025intermimic}, but translating this to robotics remains challenging. 
VideoMimic~\cite{videomimic} applies this idea to human–terrain traversal by reconstructing motions and terrains from video, but suffers from artifacts and is limited to static–scene interactions. To bridge this gap, \OmniRetarget enables natural, agile robot-object-scene interactions with high-quality reference from retargeting without manual post-processing or reward engineering.

\subsection{Data Generation for Humanoid Loco-Manipulation}
The demand for whole-body interaction data has motivated many prior works on data generation.
One approach is direct human teleoperation \cite{seo2023deep, fu2024humanplus, he2024omnih2o, ze2025twist, ben2025homie}. While it provides online feedback, teleoperation is difficult to scale: it's labor-intensive, prone to operator fatigue, and limited by the embodiment gap between human and robot kinematics. The lack of rich haptic feedback and difficulty stabilizing extreme motions (e.g., deep squats) further constrain its applicability. To address these scaling challenges, automated data augmentation has been explored, particularly for robotic manipulation. Many works leverage state-of-the-art generative models for visual \cite{zhang2024diffusion, tian2024view, chen2024rovi} and semantic \cite{mandi2022cacti, chen2023genaug, yu2023scaling} augmentations, while others rely on simple open-loop kinematic replay of base trajectories  \cite{mandlekar2023mimicgen, jiang2024dexmimicgen, garrett2024skillmimicgen} or trajectory optimization \cite{yang2025physics} in simulation. 
Despite the advances in manipulation, data augmentation for whole-body loco-manipulation remains largely unexplored. 
The closest prior work~\cite{starke2019neural} interpolates keypoints to augment objects of different shape, but it cannot deal with varied object poses either. 
\OmniRetarget directly addresses this gap.


%% file: sections/03_kinematic_retarget.tex
\section{Interaction-Preserving Motion Retargeting}
\begin{table*}[t]
\centering
\footnotesize
\renewcommand{\arraystretch}{1.15}
\setlength{\tabcolsep}{4pt}
\begin{tabularx}{\textwidth}{l *{5}{>{\centering\arraybackslash}X} l}
\toprule
\textbf{Methods} & \textbf{Hard Kinematic Constraints} & \textbf{Interaction w/ Object} & \textbf{Interaction w/ Terrain} & \textbf{Data Augmentation} & \textbf{Optimization Method} \\
\midrule
IMMA~\cite{Nakaoka2012Interaction}  & \yes & \no & \no & \no & QP \\
PHC~\cite{Luo2023PerpetualHC} & \no & \no & \no & \no & Gradient Descent \\
GMR~\cite{ze2025twist}  & \no & \no & \no & \no & Mink~\cite{Zakka_Mink_Python_inverse_2025} \\
VideoMimic~\cite{videomimic}  & Soft Penalty & \no & \yes &\no & JAX L-M \\
\midrule
\textbf{\OmniRetarget (Ours)}  & \yes & \yes & \yes &\yes & Sequential SOCP \\
\bottomrule
\end{tabularx}
\caption{Comparison of prior retargeting methods across different aspects.}
\label{tab:retargeting_comparison}
\end{table*}



\subsection{Interaction Mesh with Hard Constraints}
We leverage the interaction mesh \cite{Ho2010Spatial} to preserve spatial relationships between body parts, objects, and the environment. The interaction mesh is defined as a volumetric structure whose vertices consist of key robot or human joints together with points sampled from objects and the environment. By shrinking or stretching this mesh, we can warp human motion onto the robot while preserving relative spatial configurations and contact relationships.  

\textbf{Interaction Mesh Construction.}  
We construct the interaction mesh by applying Delaunay tetrahedralization \cite{si2005meshing} to user-defined key joint positions and randomly sampled object and environment points. To more accurately maintain contact relationships, we sample the object and environment surfaces more densely than the body joints. 

\textbf{Optimization Objectives and Constraints.}
To preserve the spatial relationships between the body parts, objects and terrains, our primary objective is to minimize the Laplacian deformation energy of the interaction meshes \cite{alexa2003differential, zhou2005large} constructed from two corresponding sets of keypoints. The source set at frame $t$, $\mathcal{P}_t^{\text{source}}$, is composed of user-defined anatomical points on the human, and points randomly sampled on the manipulated object and the environment. The target set for the retargeted motion, $\mathcal{P}_t^{\text{target}}$, consists of corresponding anatomical points on the robot, the same manipulated object and environment points. Our method is relatively robust to the precise placement of these keypoints, requiring only a semantically consistent correspondence between the human and robot (e.g., hand to hand).

The Laplacian coordinate of the $i$-th keypoint $p_{t, i} \in \mathcal{P}_t$ is defined as the difference between the point and the weighted average of its neighbors $j \in \mathcal{N}(i)$:
\begin{equation}
    L(p_{t, i}) = p_{t, i} - \sum_{j \in \mathcal{N}(i)} w_{ij} \cdot p_{t, j},
\end{equation}
where $w_{ij}$ is the normalized weight and we omit $L$'s dependence on $\{p_{t, j}\}_{j\neq i}$ in the function definition for conciseness. For all our experiments, we use uniform weights, setting $w_{ij} = 1/|\mathcal{N}(i)|$. The deformation energy measures the change in these Laplacian coordinates between the source demonstration mesh $\mathcal{P}_t^{\text{source}}$ and the retargeted mesh $\mathcal{P}_t^{\text{target}}$: 
\begin{equation} \label{eq:deformation_energy}
    E_L = \sum_{p_{t,i} ^{\text{source}} \in \mathcal{P}_t^{\text{source}}, p_{t, i}^{\text{target}} \in \mathcal{P}_t^{\text{target}}} \|L(p_{t,i} ^{\text{source}}) - L(p_{t,i}^{\text{target}})\|^2.
\end{equation}

We seek the robot configuration $q_t$, consisting of the floating base pose (quaternion and translation) and all joint angles, that minimizes this deformation energy while satisfying a set of hard kinematic constraints. The robot's keypoints are determined by its configuration $q_t$ via forward kinematics $f_i$ as $p_{t, i}^{\text{robot}}(q_t) = f_i(q_t) \in \mathcal{P}_t^{\text{target}}$. At each time step, we solve the following constrained, nonconvex program:
 
\begin{subequations}
\vspace{-0.4cm}
\label{eq:interaction_mesh_opt}
    \begin{align}
    q_t^\star = \argmin_{q_t} \; & \sum_{i} \|L(p_{t,i} ^{\text{source}})-L(p_{t,i}^{\text{target}}(q_t))\|^2 + \|q_t - q_{t-1}\|_{Q}^2 \label{eq:interaction_mesh_cost}\\
    \text{s.t.} \quad & \phi_j(q_t) \geq 0, \forall j \label{eq:non-penetration_cstr}\\
    & q_{\min} \leq q_t \leq q_{\max} \label{eq:joint_lmt}\\
    & v_{\min} \cdot dt \leq q_t - q_{t-1} \leq v_{\max} \cdot dt \label{eq:vel_lmt} \\
    & p_t^{F} = p_{t-1}^{F}, \forall \text{stance foot}, \label{eq:foot_sticking_cstr} 
\end{align}
\end{subequations}
where $Q$ is a cost matrix that encourages temporal smoothness, $\phi_j$ denotes the signed distance function for the $j$-th collision pair, $q_{\text{min}} / q_{\text{max}}$ and $v_{\text{min}} / v_{\text{max}}$ are the configuration and velocity bounds, $p_t^{F}$ denotes the foot position. A foot is considered to be in the stance phase if its horizontal velocity in the source motion (in the xy-plane) falls below a threshold of 1 cm/s. This optimization program solves for a temporally consistent robot trajectory that minimizes interaction mesh deformation, subject to hard constraints for collision avoidance \eqref{eq:non-penetration_cstr}, joint and velocity limits \eqref{eq:joint_lmt}--\eqref{eq:vel_lmt}, and preventing foot skating \eqref{eq:foot_sticking_cstr}.

We solve \eqref{eq:interaction_mesh_opt} sequentially for each timestep using a customized Sequential Quadratic Programming (SQP)-style solver. Within each iteration, the objective \eqref{eq:interaction_mesh_cost} is quadratically approximated and the hard constraints \eqref{eq:non-penetration_cstr}--\eqref{eq:foot_sticking_cstr} are linearized around the solution from the previous iteration. To ensure temporal consistency and accelerate convergence, the optimization at frame $t$ is warm-started with the optimal solution from the previous frame ${q^\star_{t-1}}$. A key challenge in this formulation is computing derivatives involving the quaternion-based floating base orientation; our implementation leverages the automatic differentiation framework in Drake \cite{drake}, which correctly handles the differential geometry of rotations on the $\mathbb{S}^3$ manifold \cite{jackson2021planning}.

\begin{figure}
\centering
\includegraphics[width=0.4\textwidth]{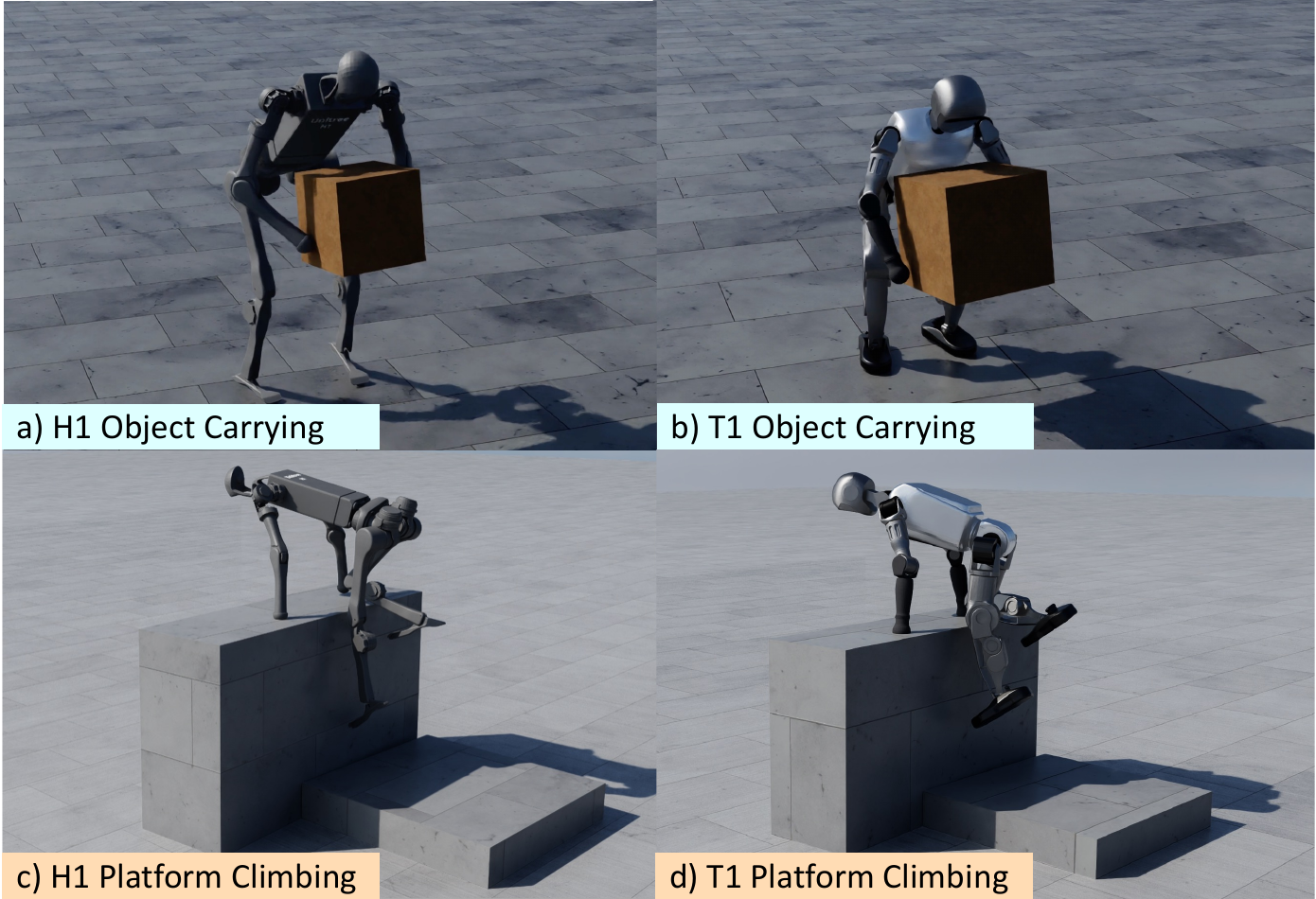}
	\caption{Cross-embodiment robot-object-terrain interaction. }
	\label{fig:cross_embodiment}
\end{figure}

Our interaction-mesh-based kinematic pipeline is highly general. It adapts to different robot embodiments, including the Unitree G1, H1, and Booster T1 (Fig.~\ref{fig:cross_embodiment}), by modifying only the keypoint correspondences in the interaction mesh and the robot’s collision model. It also supports diverse interaction types: \textbf{robot-object} interactions from the OMOMO \cite{li2023object}, \textbf{robot-terrain} interactions from in-house MoCap data, and \textbf{robot-only} motions on flat terrain from LAFAN1 \cite{harvey2020robust}.


\begin{figure*}
\centering
\includegraphics[width=\textwidth]{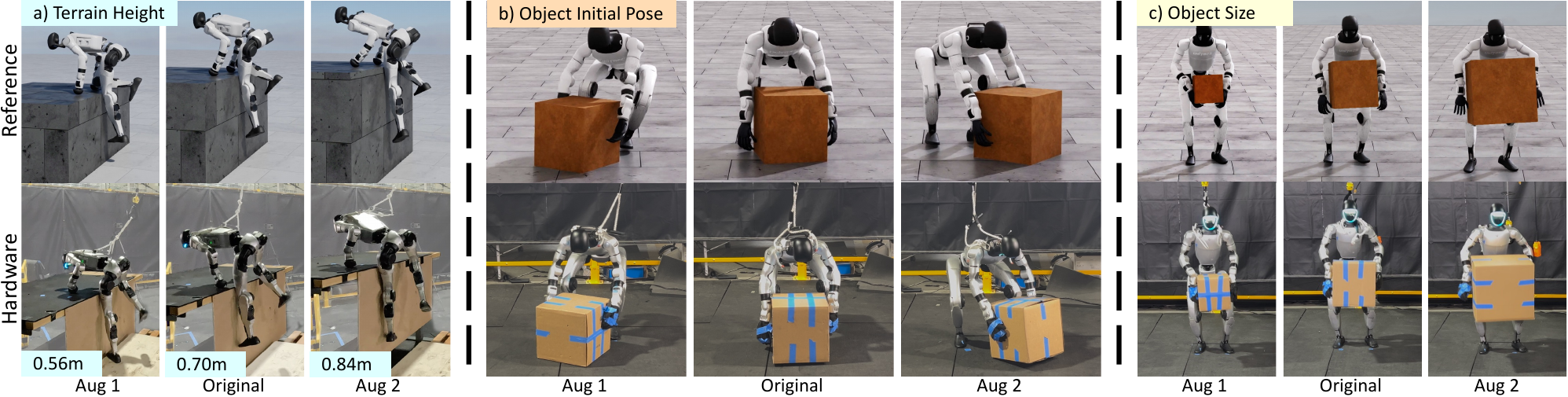}
	\caption{\label{fig:aug} \OmniRetarget generates systematic variations of (a) terrain height, (b) object initial pose, and (c) object shape from a single human demonstration, with optimized motions in simulation (top) transferring consistently to hardware (bottom). }
	\label{fig:all_aug}
\end{figure*}

\subsection{Terrain, Object Shape and Spatial Augmentation}
\label{sec:aug}
A key advantage of our framework is its capability for systematic data augmentation, which eliminates the need for collecting numerous, repetitive demonstrations with minor spatial variations. Our method can transform a single human demonstration into a rich and diverse dataset by parametrically altering object configurations, shapes, or terrain features. For each new scenario, we re-solve the optimization problem with fixed $\mathcal{P}_t^{\text{source}}$ and augmented $\mathcal{P}_t$: minimizing the interaction mesh deformation finds a new, kinematically valid robot motion $\{q_t\}$ that preserves the essential spatial and contact relationships of the original interaction.

\textbf{Robot-Object.}
We generate diverse interactions by augmenting both the object's spatial configuration and its shape. 
We apply translations and rotations to modify the object's initial pose (Fig. \ref{fig:all_aug}b) and blend the new initial pose with the original object motion via interpolation with an exponential schedule detailed in \eqref{eq:aug_obj_traj}. 
In addition, we scale the three dimensions of the object (Fig. \ref{fig:all_aug}c). A critical component of this process is constructing the interaction mesh in the object's local frame, which ensures that the robot's interacting body parts naturally follow the object's transformation (Sec. \ref{sec:im_in_obj_frame}). 

However, this alone can lead to trivial augmentations where the entire robot undergoes a rigid transformation along with the object. To generate more meaningful data diversity, we introduce cost terms and constraints that anchor parts of the robot's body to the nominal trajectory $\{\bar q_t^\star\}$. For instance, in a pick-up task, we encourage the robot to discover new upper-body coordination by penalizing lower-body deviations from the original motion:
\begin{equation}
    \|q_t - \bar q_{t}^\star\|_W,
\end{equation}
where $W$ heavily penalizes the lower-body entries, constraining the initial foot poses to match the nominal trajectory
\begin{equation}
    p_0^{F} = \bar p_{0}^{F \star} \quad \text{ for left and right feet}.
\end{equation}
These added objectives prevent the optimization from collapsing to a simple rigid transform of the nominal trajectory and instead produce genuinely new and diverse interactions.

\textbf{Robot-Terrain.}
We generate diverse terrain scenarios by scaling environmental features, such as varying the platform height and depth (Fig. \ref{fig:all_aug}a), and introducing additional constraints. For instance, to encourage stable ground contact when the terrain is elevated, we uniformly sample a grid of points on the ground surface into the interaction mesh.

%% file: sections/04_training.tex
\section{RL Training with Minimal Formulation}


Having established our method for generating high-quality kinematic references, we use RL to bridge the gap to dynamics by training a low-level policy that converts these trajectories into physically realizable actions, enabling zero-shot transfer from simulation to hardware.

Reward engineering is often the main difficulty in humanoid RL: prior works~\cite{ze2025twist, he2025asap, li2025reinforcement} rely on many ad-hoc regularizers (e.g., foot flight and contact time) to compensate for artifacts in noisy references, but tuning these terms is tedious and fragile. In contrast, BeyondMimic~\cite{liao2025beyondmimic} shows that when references are clean~\cite{unitree_lafan1_retargeting_dataset}, a minimal reward is already sufficient for high-quality tracking.
Since OmniRetarget produces such artifact-free, interaction-preserving references, we can follow this minimal formulation directly, achieving faithful tracking of dynamic interactions and zero-shot sim-to-real transfer \emph{without any hyperparameter tuning}. 

\textbf{Observations.} 
To show that high-quality reference motions provide a sufficient prior for complex tasks, we design a \emph{minimal proprioceptive} observation space, as listed below, where the agent is blind to explicit scene and object information and must follow the reference trajectory precisely. 

\begin{itemize}
    \item \emph{Reference Motion: } Reference Joint Position/Velocity, Reference Pelvis Position/Orientation Error;
    \item \emph{Proprioception: } Pelvis Linear/Angular Velocity, Joint Position/Velocity;
    \item \emph{Previous Action: } Policy action from last timestep. 
\end{itemize}

For agile motions where state estimation is unreliable, we mask out the pelvis linear position error and velocity.

\textbf{Rewards.} 
To show the benefits of high-quality reference and avoid reward tuning, we use only five reward terms: 
\begin{itemize}
    \item \emph{Body Tracking: } DeepMimic-style tracking term for body position, orientation, linear and angular velocity;
    \item \emph{Object Tracking (where applicable): } DeepMimic-style tracking term for object position and orientation;
    \item \emph{Action Rate: } Penalize rapid changes in action;
    \item \emph{Soft Joint Limit: } Penalize robot joint limit violation;
    \item \emph{Self-Collision: } Binary penalty on each body if its self-collision force exceeds $1$ N.
\end{itemize}
We use the same weights and hyperparameters from~\cite{liao2025beyondmimic} out of the box without tuning. For object tracking, we use the same hyperparameters as body tracking terms. 

\textbf{Termination.} 
We terminate training episodes with large body tracking deviations~\cite{liao2025beyondmimic}. For object loco-manipulation, episodes terminate when the object deviates more than $1.0\text{m}$ and $45$° from the reference trajectory. We only apply this criterion after the policy achieves reasonable body tracking.

\begin{figure*}
\centering
\includegraphics[width=\textwidth]{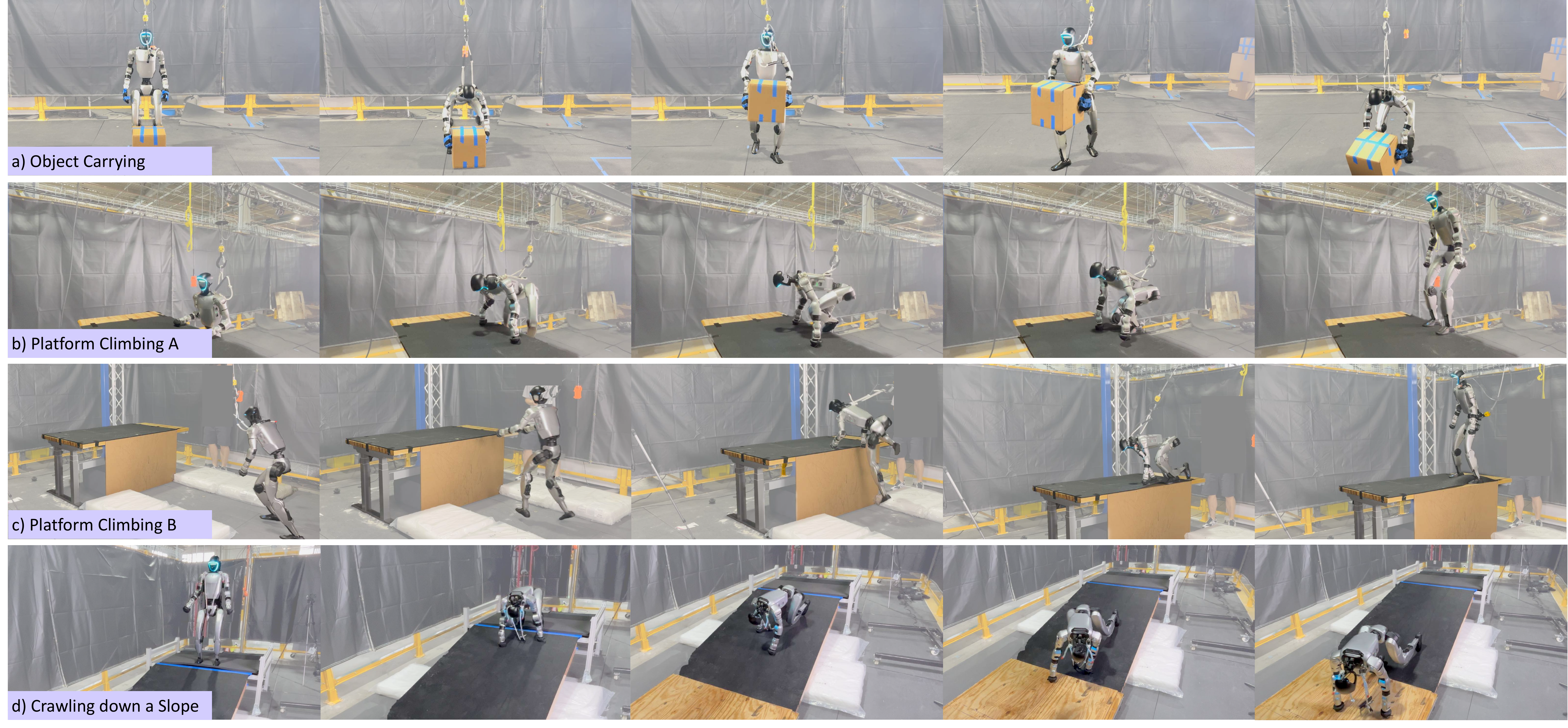}
	\caption{Additional hardware results showing diverse, agile and human-like behaviors. }
	\label{fig:results_capacity}
\end{figure*}

\textbf{Domain Randomization.}
To improve generalization across object properties for a single reference motion, we randomize the object's physical parameters: mass (0.1–2 kg), center of mass (±0.08 m), inertia (50–150\%), and shape (±10\%). 
For the robot specifically, in contrast to the many terms in prior works (e.g., random force injection (RFI), motor PD, action delay), we only apply four terms: 
\begin{itemize}
    \item \emph{Torso COM Position}: $\pm 0.025$ m in $x$, $\pm 0.05$ m in $y$, $\pm 0.075$ m in $z$;
    \item \emph{Joint default position}: $\pm 0.01$ rad;
    \item \emph{Random push}: $0.3$ m/s, $0.78$ rad/s for $(1\text{--}3)$ s;
    \item \emph{Observation noise}: $\pm 0.05$ for orientation in Rot6D, $\pm 0.5$ m/s and $\pm 0.2$ rad/s for linear and angular velocity, $\pm 0.01$ rad and $\pm 0.5$ rad/s for joint position and velocity. 
\end{itemize}

\textbf{Policy Training.}
We group similar motions for faster training. All box-moving motions share a single multi-task policy, while platform climbing uses one policy per reference.

%% file: sections/05_experiments.tex
\section{Experimental Results}
In this section, we present a comprehensive experimental validation of \OmniRetarget. We first demonstrate the breadth of complex behaviors enabled by our approach, including natural object manipulation and terrain interaction. We then provide a quantitative benchmark against state-of-the-art baselines, evaluating performance across kinematic quality metrics and downstream policy performance.
\subsection{Whole-Body Scene-Object-Interaction} 
\paragraph{Agile Loco-Manipulation}
\OmniRetarget enables RL policies to learn agile, whole-body motions for complex scene interactions and loco-manipulation in simulation, culminating in successful zero-shot sim-to-real transfer to hardware. Shown in Fig.~\ref{fig:results_capacity}, policies trained on our data reproduce a diverse range of expressive behaviors on a Unitree G1 humanoid, including natural box-carrying motions retargeted from the OMOMO dataset, dynamically climbing a $0.9$m-high platform ($70$\% of the robot's height), and crawling over slopes, showing clean and accurate contact sequences. 

To showcase the full capabilities of our framework, we present a long-horizon, dynamic sequence inspired by the Boston Dynamics Atlas tool-use demo \cite{BostonDynamics2023Atlas}. Visualized in Fig.~\ref{fig:flagship_demo}, our retargeted data enables the robot to carry a $4.6$ kg chair to a platform, use it as a stepstone to climb up, and then leap off, performing a parkour-style roll to absorb the landing impact. This 30-second, complex, multi-stage task highlights \OmniRetarget's ability to produce precise and versatile reference motions, pushing the boundaries of what is possible for humanoids learning agile, human-like behaviors.

\begin{figure*}
\centering
\includegraphics[width=\textwidth]{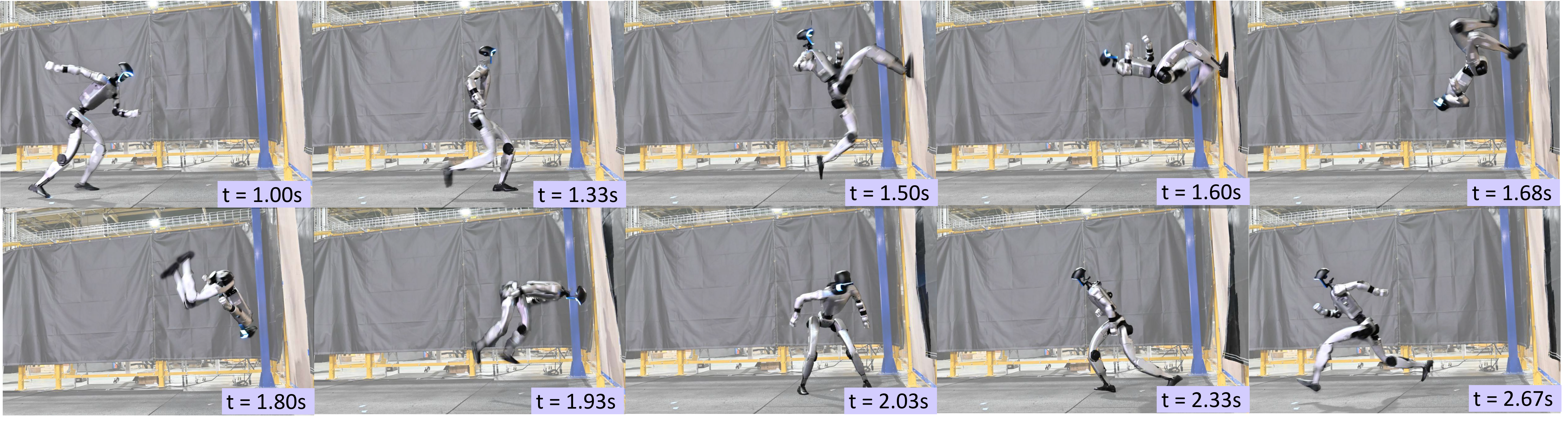}
	\caption{Hardware results showing a high-dynamic wall-flip motion. The robot reaches a maximum linear velocity of $3.5$ m/s and a peak angular velocity of $15$ rad/s. }
	\label{fig:wallflip}
\end{figure*}

We additionally showcase a high-dynamic wall-flip motion\footnote{The motion is acquired from \url{https://actorcore.reallusion.com/3d-motion?asset=parkour-tic-tac-backflip}
. An IMU capable of measuring angular rates above $15$ rad/s is required for this motion.} in Fig.~\ref{fig:wallflip}. The robot completes the full flip in approximately $0.5$ second, reaching a peak angular velocity of $15$ rad/s.
Unlike the human foot, which can flex at the arch to maintain contact and generate friction, the robot foot is rigid. As a result, it must align more closely to the wall to achieve sufficient contact area and friction.
To account for this physical difference and give RL more freedom to learn this skill, we relaxed the termination condition during RL training by increasing the end-effector position error threshold to $0.5$ meter (compared to $0.25$ meter used in other motions) and removed the foot joint orientation tracking term from the reward function. All other components of the tracking objective remain consistent with other motions.
The trained policy is robust and achieves a $5/5$ success rate in our real-world experiments.

\paragraph{Sim-to-real with Augmented Data} 
We show that the augmented motions from our pipeline can be used for training and deployment effectively. As shown in Fig.~\ref{fig:aug}, the interaction mesh formulation allows \OmniRetarget to generalize a single nominal motion into box-picking across shapes and positions, as well as platform climbing at different heights. Notably, these augmented motions transfer to hardware without reward tuning, effectively expanding the repertoire of scenes and behaviors we can achieve in real.

In comparison, relying solely on domain randomization--which perturbs object shapes and poses only during training--performs poorly under our RL formulation, as the policies struggle to explore far beyond the nominal reference. Policies trained on our augmented data instead yield reliable success (see video for comparison). Admittedly, additional reward engineering could help, but it contradicts our minimal design goal. Quantitatively, training and evaluating on the full augmented dataset achieves a $79.1\%$ success rate, comparable to $82.2\%$ when evaluating on nominal motions only, showing that kinematics augmentation substantially enlarges coverage without significant performance degradation.


\begin{table*}[tb]
\centering 
\begin{tabular}{lcccccc} 
\toprule 
& \multicolumn{2}{c}{\textbf{Penetration}} & \multicolumn{2}{c}{\textbf{Foot Skating}} & \textbf{Contact Preservation} &
\textbf{Downstream RL Policy}\\ 
\cmidrule(lr){2-3} \cmidrule(lr){4-5} \cmidrule(lr){6-6} \cmidrule(lr){7-7} 
\textbf{Method} & Duration $\downarrow$ & Max Depth (cm) $\downarrow$ & Duration $\downarrow$ & Max Vel. (cm/s) $\downarrow$ & Duration $\uparrow$ & Success Rate $\uparrow$\\ 
\midrule
\multicolumn{6}{l}{\textit{Robot-Object Interaction (Retargeting from the OMOMO Dataset)}} \\
\midrule 
PHC~\cite{Luo2023PerpetualHC} & 0.68 $\pm$ 0.21 & 5.11 $\pm$ 3.09 & 0.05 $\pm$ 0.05 & 1.40 $\pm$ 0.80 & 0.96 $\pm$ 0.09 & 71.28\% $\pm$ 22.55\%\\ 
GMR~\cite{ze2025twist} & 0.83 $\pm$ 0.14 & 8.50 $\pm$ 3.94 & 0.02 $\pm$ 0.01 & 1.46 $\pm$ 0.45 & \textbf{0.99 $\pm$ 0.04} & 50.83 \% $\pm$ 23.89\% \\  
VideoMimic~\cite{videomimic} & 0.60 $\pm$ 0.27 & 7.48 $\pm$ 4.95 & 0.12 $\pm$ 0.07 & 1.50 $\pm$ 0.70 & 0.77 $\pm$ 0.25 & 3.85\% $\pm$ 8.41\%\\ 
\OmniRetarget & \textbf{0.00 $\pm$ 0.01} & \textbf{1.34 $\pm$ 0.34} & \textbf{0} & \textbf{0} & 0.96 $\pm$ 0.09 & \textbf{82.20\% $\pm$ 9.74\%}\\
\midrule
\multicolumn{6}{l}{\textit{Robot-Terrain Interaction (Retargeting from the In-House MoCap Dataset)}} \\
\midrule
PHC & 0.66 $\pm$ 0.36 & 7.74 $\pm$ 4.53 & 0.15 $\pm$ 0.04 & 2.03 $\pm$ 1.83 & 0.45 $\pm$ 0.28 & 52.63\% $\pm$ 49.93\% \\ 
GMR & 0.91 $\pm$ 0.16 & 5.72 $\pm$ 3.84 & 0.04 $\pm$ 0.05 & 1.75 $\pm$ 3.01 & 0.67 $\pm$ 0.26 & 78.94\% $\pm$ 40.77\% \\ 
VideoMimic & 0.83 $\pm$ 0.11 & 5.97 $\pm$ 3.58 & 0.14 $\pm$ 0.05 & 1.85 $\pm$ 1.38  & 0.47 $\pm$ 0.25 & 51.75\% $\pm$ 49.23\% \\ 
\OmniRetarget & \textbf{0.01 $\pm$ 0.02} & \textbf{1.37 $\pm$ 0.18} & \textbf{0} & \textbf{0} & \textbf{0.72 $\pm$ 0.19} & \textbf{94.73\% $\pm$ 22.33\%}\\ 
\midrule
\multicolumn{6}{l}{\textit{Robot-Only (Retargeting from the LAFAN1 Dataset)}} \\
\midrule
Unitree~\cite{unitree_lafan1_retargeting_dataset} & 0.09 $\pm$ 0.13 & 3.22 $\pm$ 2.64 & 0.06 $\pm$ 0.03 & 1.46 $\pm$ 0.01 & N/A & \textbf{100\%} \\
\OmniRetarget & \textbf{0.00 $\pm$ 0.00} & \textbf{1.07 $\pm$ 0.00} & \textbf{0} & \textbf{0} & N/A & \textbf{100\%} \\
\bottomrule 
\end{tabular}
\caption{\label{tab:kinematic_quality} Quantitative comparison of kinematic retargeting quality and downstream RL performances.}
\vspace{-0.4cm}
\end{table*}

\subsection{Benchmark Against Prior Retargeting Pipelines}
\begin{figure}
\centering
\includegraphics[width=0.48\textwidth]{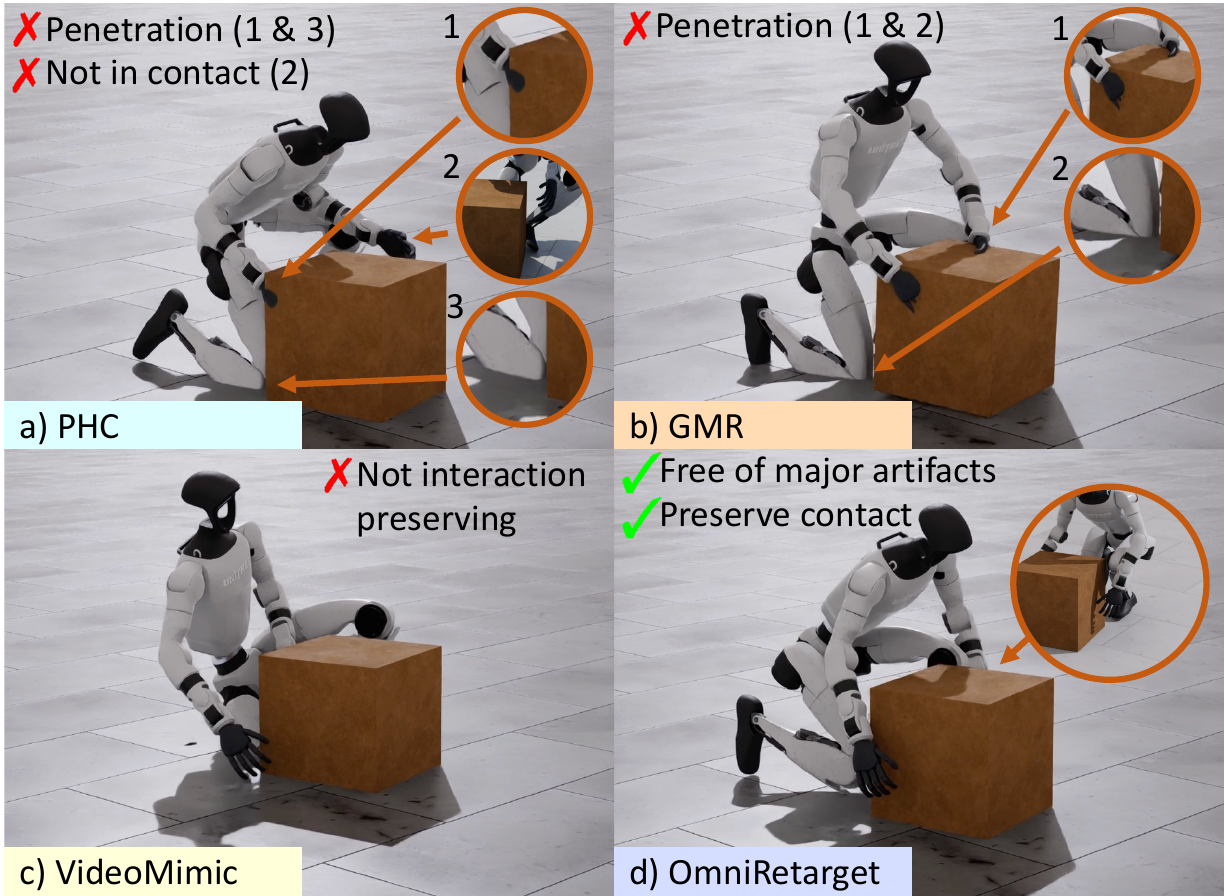}
	\caption{\label{fig:artifact}Artifacts resulting from the retargeting baselines. 
    }
	\label{fig:baseline_artifacts}
    \vspace*{-0.4cm}
\end{figure}

We compare \OmniRetarget against several widely-used open-source retargeting baselines\footnote{Baseline performance may depend on their hyperparameters. We initialized from the default settings in their public codes, and further improved to ensure consistent performance across different tasks.}: PHC~\cite{Luo2023PerpetualHC}, GMR~\cite{ze2025twist} and VideoMimic~\cite{videomimic}. The generated dataset including 2.78 hours of box carrying in OMOMO, 1 hour of in-house MoCap and 4.6 hours of LAFAN1 will be open-sourced. 
\paragraph{Kinematics Quality} We evaluate the kinematic quality of retargeted motions on a Unitree G1 with three criteria:
\begin{enumerate}
    \item \textbf{Penetration}: Measured by the time duration (normalized by the trajectory length) and maximum depth of intersections between the robot, objects, and terrain.
    \item \textbf{Foot Skating}: Quantified by the time duration (normalized by the total desired foot sticking length) and maximum skating velocity of a stance foot.
    \item \textbf{Contact Preservation}: Quantified by the time duration (normalized by the desired contact length). For \textit{robot-object} tasks, we measure hand-object contact. For \textit{robot-terrain} tasks, we measure contact between the robot's hands, toes, and heels with the terrain surface.
\end{enumerate}
As illustrated in Tab.~\ref{tab:kinematic_quality}, \OmniRetarget significantly outperforms all baselines across most kinematic metrics. While \OmniRetarget occasionally incurs minor penetration due to the linearization of constraints \eqref{eq:non-penetration_cstr} in the sequential SOCP solver, the violations are minimal and can be efficiently fixed by RL. 
GMR achieves the highest contact preservation score for robot-object interaction tasks; however, this outcome largely reflects its keypoint-matching objective. In practice, scaling human hand keypoints to the robot’s size often drives the robot’s hands inside the object, leading to substantial penetration errors (Fig. \ref{fig:baseline_artifacts}b). Overall, all baselines exhibit significant penetration and foot skating (Fig. \ref{fig:baseline_artifacts}), degrading the downstream RL performance, as discussed next.

For a direct comparison on pure locomotion, we retarget motions from the LAFAN1 MoCap dataset \cite{harvey2020robust} and benchmark them against the publicly available Unitree LAFAN1 retargeted dataset \cite{unitree_lafan1_retargeting_dataset}. This serves as a strong baseline, as it is widely considered a high-quality data source for RL-based locomotion training~\cite{liao2025beyondmimic}. Table~\ref{tab:kinematic_quality} shows that \OmniRetarget's motions exhibit fewer physical artifacts, achieving better satisfaction of hard constraints.

\paragraph{Downstream RL Performance}
\label{sec:downstream-rl}
A central observation from prior works~\cite{liao2025beyondmimic, zhang2025hub} is that the quality of retargeted motions strongly influences the performance of downstream RL. To verify this, we select 39 challenging motions for \OmniRetarget and baselines, and train RL policies using identical hyperparameters from~\cite{liao2025beyondmimic} without manual tuning. We evaluate the policies in simulation, and success is measured by training termination criteria.

Shown in Tab.~\ref{tab:kinematic_quality}, retargeting quality directly impacts RL success rates. \OmniRetarget consistently achieves the highest performance across tasks, exceeding baselines by over 10\% with lower variance, which indicates more stable learning across different motions. PHC performs better than GMR in object manipulation, likely due to lower penetration with sufficient contact preservation, but worse in terrain interaction, where its contact preservation drops by nearly 50\%. Specifically for terrain interaction, we see that contact preservation is directly proportional to the success rate. These results suggest that both contact preservation and penetration reduction are critical for generalizing RL policies across diverse tasks, and \OmniRetarget shows strength in both.

VideoMimic shows the weakest interaction preservation among all baselines (Fig.~\ref{fig:artifact}c), likely due to its collision avoidance soft cost conflicting with the keypoint matching cost. 
This is compounded by its coarse collision model originally designed for heightmaps, which is ill-suited for precise loco-manipulation. Consequently, while its terrain-interaction results are comparable to PHC, its performance on object manipulation is poor. Although this could be partially attributed to the tuning of its soft penalties, \OmniRetarget demonstrates that a hard-constraint formulation avoids such sensitivities altogether. 

%% file: sections/06_conclusion.tex
\section{Conclusion}

In this work, we tackled a key data bottleneck caused by a lack of high-quality, interaction-aware retargeting pipeline in humanoid whole-body loco-manipulation. 
We introduced \OmniRetarget, a unified, interaction-preserving data generation engine that leverages an interaction mesh within a single constrained optimization. Our experiments showed that \OmniRetarget significantly outperforms prior methods in kinematic quality, producing a diverse set of artifact-free trajectories from single demonstrations. 
This high-quality data enabled a proprioceptive RL policy, trained with minimal formulation, to achieve long-horizon dynamic skills on a physical humanoid via zero-shot sim-to-real transfer.

Ultimately, \OmniRetarget demonstrates a paradigm shift from patching lower-quality reference motions with complex reward engineering to solving the problem at its source with principled data generation. While our current frame-by-frame optimization is mostly effective, future work could explore jointly optimizing the entire trajectory to enhance the framework's robustness to noisier motion sources, such as video data, or learning autonomous visuomotor policies. By open-sourcing our complete framework and the large-scale dataset of retargeted trajectories, we hope to accelerate progress towards more agile, capable, and versatile humanoid robots. \looseness=-1

%% file: sections/07_appendix.tex
\section*{APPENDIX}
\subsection{Different Sources of Human Motion Data}
Human motion datasets contain rich pose and shape information, but they differ both in data format and in the physical attributes (e.g., height, body proportions) of the demonstrators. To make them compatible across different sources and suitable for retargeting, we need to convert these inputs into a consistent representation, typically a time series of global 3D keypoint positions $\{p^\text{source}_{0:T, i}\}$. This process must account for differences between human demonstrators and the target robot.

The datasets used in this work represent two common formats:
\begin{itemize}
    \item \textbf{Parametric Human Models}: The OMOMO dataset uses the SMPL format \cite{SMPL:2015}, a parametric model representing human body shape and pose-dependent variations using shape ($\beta$) and pose ($q$) parameters.
    \item \textbf{Skeleton Hierarchy}: Both our in-house MoCap data and the LAFAN1 dataset utilize the skeleton hierarchy defined in the BVH format.
\end{itemize}
Different retargeting pipelines use different strategies to handle these formats. We detail these preprocessing steps below, denoting the human demonstrator's pose as $q^\text{demo}_{t}$, the SMPL forward model for the $i$-th keypoint as $M_i$, the original demonstrator shape as $\beta^{\text{demo}}$, and the demonstrator's $i$-th keypoint position as $p^\text{demo}_{t,i}$.

\subsubsection{SMPL Data}
To handle data from parametric models like SMPL, methods typically follow one of two strategies: fitting the model to the robot's morphology or directly scaling the human's keypoints.
\begin{algorithm}
\caption{Fit SMPL Shape (PHC)}
\begin{algorithmic}[1]\label{alg:phc_fit_smpl_shape}
\REQUIRE SMPL model $M$, robot urdf with forward kinematics $f$, $\bar q^\text{smpl} = 0_{n_s}, \bar q^\text{robot} = 0_{n_x}$
\ENSURE scaling factors $\alpha, \beta$
\STATE $\alpha, \beta \leftarrow 1, 0_{10}$
\FOR{$\text{iter} = 1, \ldots, \text{max\_iter}$}
    \STATE $L(\alpha, \beta) = \sum_i \left\|(f_i(\bar q^\text{robot}) - \alpha \cdot M_i(\bar q^\text{smpl}; \beta)\right\|^2$
    \STATE $\alpha \leftarrow \alpha - \eta_{\alpha} \cdot \nabla_{\alpha} L$
    \STATE $\beta \leftarrow \beta - \eta_{\beta} \cdot \nabla_{\beta} L$
\ENDFOR
\end{algorithmic}
\end{algorithm}
\paragraph{Model Fitting (PHC, VideoMimic)} This strategy fits a scaled SMPL model to the robot's morphology. PHC first optimizes for an overall scaling factor $\alpha$, and a set of SMPL shape parameters $\beta$ that best match the robot's link length in a canonical T-pose, as detailed in Alg. \ref{alg:phc_fit_smpl_shape}. The final source keypoint positions are then generated from this fitted model:\looseness=-1
\begin{equation}
p^\text{source}_{t,i} = \alpha \cdot M_i(q^\text{demo}_t; \beta).
\end{equation}
VideoMimic adopts a similar philosophy but integrates the scaling directly into its main retargeting optimization, solving for per-link scale factors jointly with the robot's motion.

\paragraph{Direct Scaling (GMR \& \OmniRetarget)}
In contrast, GMR and \OmniRetarget use a more direct approach. They generate keypoints from the human's \emph{original} SMPL parameters $\beta^{\text{demo}}$ and then scale them to the robot's proportions. 
Both methods support detailed morphological adaptation via per-bone scaling factors based on corresponding human-robot link lengths. For simplicity in this work, however, we adopt a single global scaling factor $\alpha$, set to the robot-to-human height ratio:

\begin{equation}\label{eq:gmr_scaling}
    p^\text{source}_{t,i} = \alpha \cdot M_i(q^\text{demo}_t; \beta^{\text{demo}}), \alpha=\frac{h_{\text{robot}}}{h_{\text{demo}}}.
\end{equation}

\begin{table*}[tb]
\centering
\begin{tabular}{@{}lllll@{}}
\toprule
\textbf{Method} & \textbf{Optimization Type} & \textbf{Primary Objective} & \textbf{Preprocessing} & \textbf{Data Formats} \\ \midrule
PHC & Trajectory-wise Optimization & Keypoint Position Matching & Model Fitting & SMPL \\
GMR & Per-Frame Optimization & Keypoint Position \& Orientation Matching  & Direct Scaling & SMPL, BVH  \\
VideoMimic & Trajectory-wise Optimization & Pairwise Distance \& Orientation Preservation & Model Fitting & SMPL \\
IMMA & Multi-Stage Trajectory-wise Optimization & Interaction Mesh Deformation + IK & Unknown & Unknown \\ \midrule
\textbf{\OmniRetarget} & Per-Frame Optimization & Interaction Mesh Deformation & Direct Scaling & SMPL, BVH   \\ \bottomrule
\end{tabular} 
\caption{\label{tab:method_comparison}Comparison of different retargeting methodologies}
\end{table*}

\subsubsection{Skeleton Hierarchy Data}
For formats like BVH, keypoint positions are derived from the skeleton's forward kinematics $f^\text{skeleton}$. This data is then typically scaled to the robot's size using the height ratio:
\begin{equation}
    p^\text{source}_{t, i} = \frac{h_{\text{robot}}}{h_{\text{demo}}} \cdot f_i^\text{skeleton}(q^\text{demo}_t).
\end{equation}

A key distinction among methods is their data compatibility. While GMR and \OmniRetarget are designed to process both parametric model data and raw skeleton hierarchies, frameworks like PHC and VideoMimic are primarily designed for SMPL data. Fitting other data formats to the SMPL format is yet another tedious process. 

\subsection{Different Kinematic Retargeting Formulations}
Once human motion is preprocessed into a series of source keypoint positions $\{p^\text{source}_{0:T, i}\}$, different methods formulate the retargeting problem in distinct ways. As summarized in Tab.~\ref{tab:method_comparison}, these approaches vary in their optimization strategy and objectives. The following sections detail the mathematical formulation of each baseline method and our proposed approach, \OmniRetarget.
\begin{algorithm}
\caption{Retarget Robot Motion (PHC)}
\begin{algorithmic}[1]\label{alg:phc_fit_robot_motion}
\REQUIRE Robot urdf, source keypoint positions $\{p^\text{source}_{0:T, i}\}$
\ENSURE $q_{0:T}$
\STATE $q_{0:T} \leftarrow [0_{n_x}]_T$
\FOR{$\text{iter} = 1, \ldots, \text{max\_iter}$}
    \STATE $\mathcal{L}(q_{0:T}) = \sum_{t=0}^T  \sum_i \left\|f_i(q_t) - p^\text{source}_{t, i}\right\|^2$
    \STATE $q_{0:T} \leftarrow \text{clamp}(q_{0:T} - \nabla_{q_{0:T}} \mathcal{L}(q_{0:T}), q_{\min}, q_{\max})$
\ENDFOR
\end{algorithmic}
\end{algorithm}

\subsubsection{PHC}
PHC formulates retargeting as a large-scale trajectory-wise optimization problem. It applies gradient descent to minimize the error between the source keypoint positions and the robot's keypoint positions over the entire trajectory, as shown in Alg.~\ref{alg:phc_fit_robot_motion}. \looseness=-1

\subsubsection{GMR}
GMR performs retargeting by solving an inverse kinematics (IK) problem at each frame (\ref{alg:gmr_retarget_robot_motion}).
At each timestep, GMR finds the robot configuration $q_t$ that matches the source keypoint positions and orientations via the following optimization program:
\begin{equation} \label{eq:gmr_qp}
    \begin{aligned}
        q_t^\star = \argmin_{q_t} & \sum_i \left\|f_i^p(q_t) - p^\text{source}_{t, i}\right\|^2 + \left\|f_i^\theta(q_t) - \theta^\text{source}_{t, i}\right\|^2 \\
      \text{s.t. } & q_{\min} \leq q_t \leq q_{\max},
    \end{aligned}
\end{equation}
where $f_i^p$ and $f_i^\theta$ are the robot forward kinematics for the $i$-th keypoint's position and orientation, respectively. Leveraging the mink \cite{Zakka_Mink_Python_inverse_2025} library, GMR solves this program in a Sequential Quadratic Programming fashion.
\begin{algorithm}
\caption{Retarget Robot Motion (GMR)}
\begin{algorithmic}[1]\label{alg:gmr_retarget_robot_motion}
\REQUIRE Robot urdf, source keypoint positions $\{p^\text{source}_{0:T, i}\}$ and orientations $\{\theta^\text{source}_{0:T, i}\}$
\ENSURE $q_{0:T}$
\FOR{$t = 0, \ldots, T$}
    \STATE $q_{t} \leftarrow$ Solve IK \eqref{eq:gmr_qp}
\ENDFOR
\end{algorithmic}
\end{algorithm}

\subsubsection{VideoMimic}

VideoMimic jointly optimizes for the robot motion $q_{0:T}$ and SMPL per-link scaling factor $\beta$ over the entire trajectory. The primary objective is to preserve the scaled pairwise distance and orientation between each keypoint pair $(i, j)$:
\begin{equation}
    \mathcal{L}_{\text{pairwise}} = \sum_{t, i \in \mathcal{N}(j)} \|\beta_{ij} \cdot (p^\text{demo}_{t,i} -  p^\text{demo}_{t,j}) - (f_i(q_t) - f_j(q_t))\|_2^2,
\end{equation}
with soft penalties on foot contact matching $\mathcal{L}_{\text{contact}}$, foot skating $\mathcal{L}_{\text{skating}}$, collision $\mathcal{L}_{\text{collision}}$, joint limits $\mathcal{L}_{\text{joint}}$ and temporal smoothness $\mathcal{L}_{\text{smooth}}$:
\begin{equation} \label{eq:videomimic_opt}
    \begin{aligned}
        q_{0:T}^\star, \beta^\star = & \argmin_{q_{0:T}, \beta} \quad  \mathcal{L}_{\text{pairwise}} + \lambda_c \cdot \mathcal{L}_{\text{contact}} + \lambda_s \cdot \mathcal{L}_{\text{skate}} + \\
        & \lambda_{cl} \cdot \mathcal{L}_{\text{collision}} + \lambda_j \cdot \mathcal{L}_{\text{joint}} + \lambda_{sm} \cdot \mathcal{L}_{\text{smooth}} + \ldots
    \end{aligned}
\end{equation}
\begin{algorithm}
\caption{Retarget Robot Motion (VideoMimic)}\label{alg:videomimic_retarget}
\begin{algorithmic}[1]
\REQUIRE Robot urdf, demonstrator's original keypoint positions $\{p^\text{demo}_{0:T, i}\}$
\ENSURE $q_{0:T}, \beta$ $\leftarrow$ Solve \eqref{eq:videomimic_opt} for the entire trajectory
\end{algorithmic}
\end{algorithm}

\subsubsection{IMMA Multi-stage Optimization}
IMMA relies on a complex, multi-stage pipeline: first, it optimizes the intermediate robot keypoint positions to warp the interaction mesh from the human to the robot with minimal deformation by solving the following program
\begin{equation}
    \begin{aligned}
    p_{t,i}^\star = \argmin_{p_{t_i}} \quad & \sum_{i} \|L(p_{t,i} ^{\text{source}})-L(p_{t,i})\|^2 \\
    \text{s.t.} \quad & \phi_j(q_t) \geq 0, \forall j \\
    & \|p_{t,i} - p_{t,j}\|_2 = l_{ij}, \forall \text{bone} \\
    & p_t^{F} = p_{t-1}^{F}, \forall \text{stance foot}, 
\end{aligned}
\end{equation}
where $l_{ij}$ is the bone length between the $i$-th and $j$-th joints. Then, it solves a separate IK problem to recover joint angles that best match the intermediate keypoints: 
\begin{equation}
    \begin{aligned}
        q_t^\star = \argmin_{q_t} \sum_i \|f_i(q_t) - p_{t,i}^\star \|_2^2.
    \end{aligned}
\end{equation}
In later stages, additional hard constraints on the feet and waist are imposed to prevent foot slipping and ensure dynamic balancing. This sequential and fragmented approach produces dynamically consistent motions but fails to consider crucial kinematic constraints like joint and velocity limits.

\begin{algorithm}
\caption{Retarget Robot Motion (\OmniRetarget)}
\begin{algorithmic}[1]\label{alg:omniretarget_retarget_robot_motion}
\REQUIRE Robot urdf, source keypoint positions $\{p^\text{source}_{0:T, i}\}$
\ENSURE $q_{0:T}$
\FOR{$t = 0, \ldots, T$}
    \STATE $q_{t} \leftarrow$ Solve interaction mesh optimization \eqref{eq:interaction_mesh_opt}
\ENDFOR
\end{algorithmic}
\end{algorithm}

\subsubsection{\OmniRetarget}
\OmniRetarget, as outlined in Alg. \ref{alg:omniretarget_retarget_robot_motion}, operates frame-by-frame by minimizing the Laplacian deformation of the interaction meshes. The core objective \eqref{eq:interaction_mesh_cost} is flexible and can be augmented with task-specific costs, such as the orientation matching term from GMR, providing a unified and extensible framework for motion retargeting.

\subsection{Data Augmentation Details}
\subsubsection{Augmented Object Trajectory}\label{sec:aug_obj_traj}
To generate a perturbed object trajectory, we introduce a transient offset that decays exponentially over time. Let the original trajectory be denoted by $(p_{obj}(t), \theta_{obj}(t))$. We define an initial positional offset $\Delta p_{obj}$ and rotational offset $\Delta \theta_{obj}$ that are applied at the onset of object motion, $t_m$. The augmented trajectory, $(\tilde{p}_{obj}(t), \tilde{\theta}_{obj}(t))$, is then formulated as:

\begin{subequations}\label{eq:aug_obj_traj}
\begin{align}
\tilde{p}_{obj}(t) &=
\begin{cases}
\Delta p_{obj} + p_{obj}(0) & \text{if } t < t_m \\
\Delta p_{obj}\, e^{-(t-t_m)/\tau_p} + p_{obj}(t)& \text{if } t \ge t_m
\end{cases}
\\[0.75em]
\tilde{\theta}_{obj}(t) &=
\begin{cases}
\Delta \theta_{obj} \oplus \theta_{obj}(0) & \text{if } t < t_m \\
\Delta \theta_{obj}\, e^{-(t-t_m)/\tau_{\theta}} \oplus \theta_{obj}(t) & \text{if } t \ge t_m
\end{cases}
\end{align}
\end{subequations}
where $\tau_p$ and $\tau_\theta$ are time constants governing the rate of decay for the translational and rotational perturbations, respectively. The $\oplus$ operator denotes composition for orientations (e.g., quaternion multiplication).

\subsubsection{Interaction Mesh Construction in Object Frame} \label{sec:im_in_obj_frame}
\begin{figure}
\centering
\includegraphics[width=0.20\textwidth]{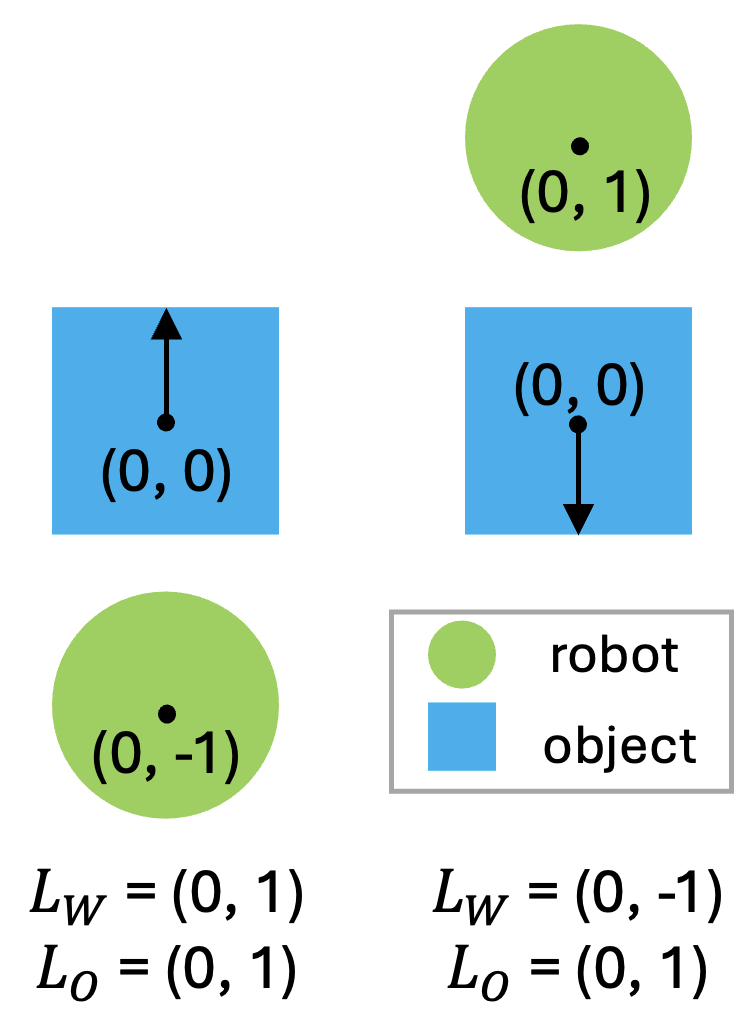}
	\caption{The Laplacian coorinate should stay the same when the object rotates \ang{180}. }
	\label{fig:obj_frame_im}
    \vspace*{-0.4cm}
\end{figure}

For robot-object interactions, it is crucial to construct the interaction mesh in the object's local coordinate frame. This ensures that the Laplacian coordinates, which encode relative spatial relationships, are invariant to the object's global rotation and translation. As illustrated in Fig. \ref{fig:obj_frame_im}, when the object rotates by \ang{180} (indicated by the black arrow), the Laplacian coordinate of the object in the world frame $L_W$ changes from $(0, 1)$ to $(0, -1)$, while the Laplacian coordinate calculated in the object frame $L_O$ remains constant. Using object-frame coordinates is therefore essential for preserving the intended interaction geometry during object spatial transformation.

\begin{figure}
\centering
\includegraphics[width=0.23\textwidth]{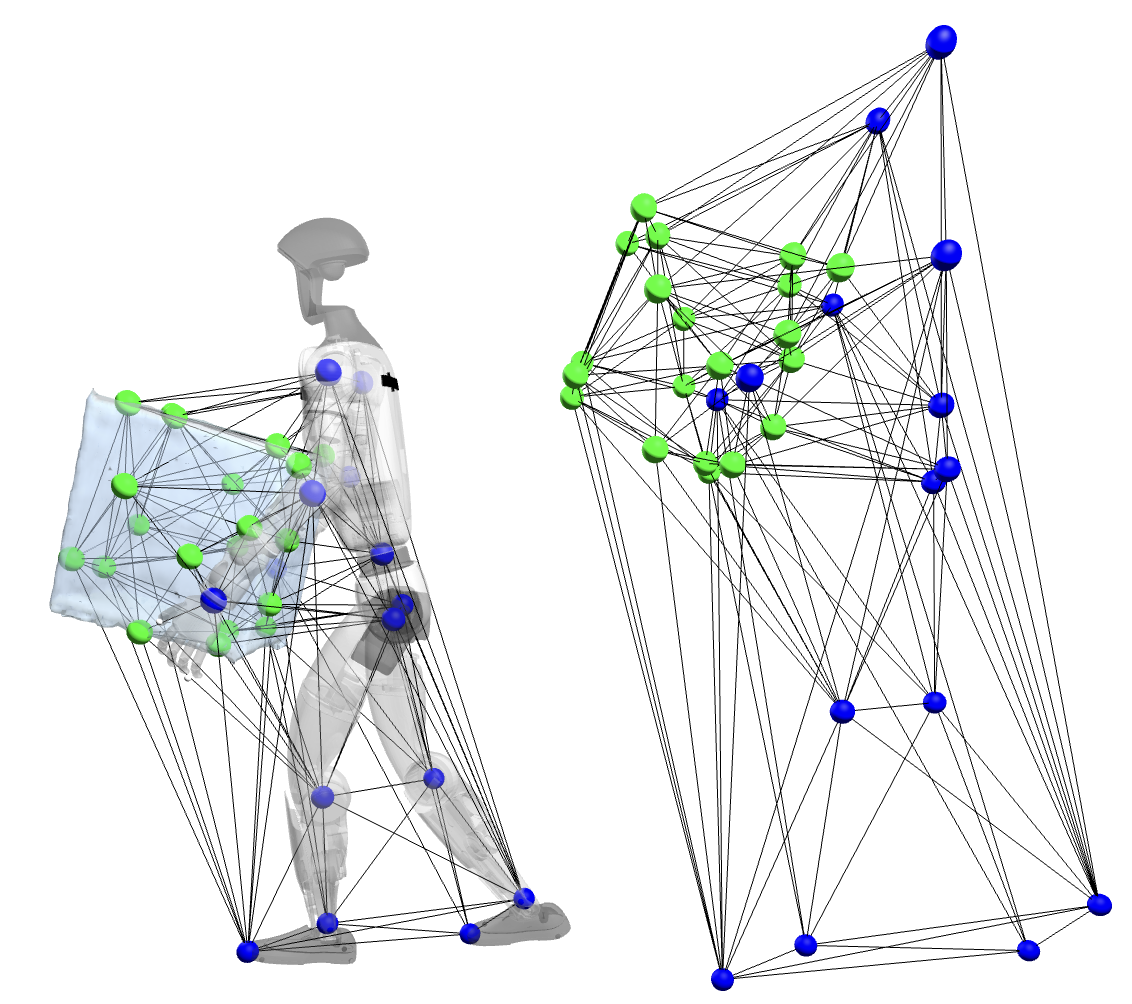}
	\caption{The actual target (left) and source (right) interaction meshes used for optimization.}
	\label{fig:interaction_mesh}
    \vspace*{-0.4cm}
\end{figure}
\subsection{Sequential SOCP Details}
At each time step $t$, we iteratively solve a Second-Order Cone Program (SOCP) for the optimal change in the robot's configuration $dq$ util convergence for up to 10 iterations. For conciseness, we omit the time index $t$ for variables within the Sequential SOCP loop. 

The optimization finds the increment $dq_n$ at the $n$-th iteration. The configuration is updated using this increment, starting from the previous time step's solution ($\bar q_0 = q_{t-1}^\star$):
$$\bar q_{n+1} = \bar q_n + dq_n^\star.$$
The optimal increment $dq_n^\star$ is the solution to the following SOCP, which is linearized around the current iterate  
$\bar q_n$:
\begin{subequations}
\begin{align}
    dq_n^\star = \argmin_{dq_n} \quad & \|L^{\text{source}} - (J_L^n\cdot dq_n +\bar L_n^{\text{target}})\|^2 \\
    & + \|\bar{q}_n + dq_n - q_{t-1}\|_{Q}^2 \\
    \text{s.t.} \quad & J_j^n \cdot dq_n + \phi_j(\bar{q}_n) \geq 0, \forall j \\
    & q_{\min} \leq \bar{q}_n + dq_n \leq q_{\max} \\
    & v_{\min} \cdot dt \leq \bar{q}_n + dq_n - q_{t-1} \leq v_{\max} \cdot dt \\
    & p_t^{F}(\bar{q}_n) + J_{F}^n \cdot dq_n = p_{t-1}^{F}, \forall \text{stance foot} \\
    & \|dq_n\|_2 \leq \epsilon, \label{eq:trust_region_cstr}
\end{align}
\end{subequations}
where 
\begin{itemize}
    \item $L^{\text{source}} = vec(\{L(p_{t,i}^{\text{source}})\})$
    \item $L^{\text{target}}(q) = vec(\{L(p_{t,i}^{\text{target}}(q))\})$
    \item $\bar L_n^{\text{target}} = vec(\{L( p_{t,i}^{\text{target}}(\bar q_n))\})$
    \item $J_L^n = \partial L^\text{target}/\partial q|_{q=\bar q_n}$
    \item $J_j^n = \partial \phi_j/\partial q|_{q=\bar q_n}$
    \item $J_F^n = \partial p_t^F /\partial q|_{q=\bar q_n}$.
\end{itemize}
The second-order cone constraint \eqref{eq:trust_region_cstr} is a trust region constraint with radius $\epsilon$ (we use $\epsilon=0.2$) that keeps the step size small, ensuring the linear approximations remain valid.

\subsection{Downstream RL Evaluation Breakdown}
\begin{figure}
\centering
\includegraphics[width=0.49\textwidth]{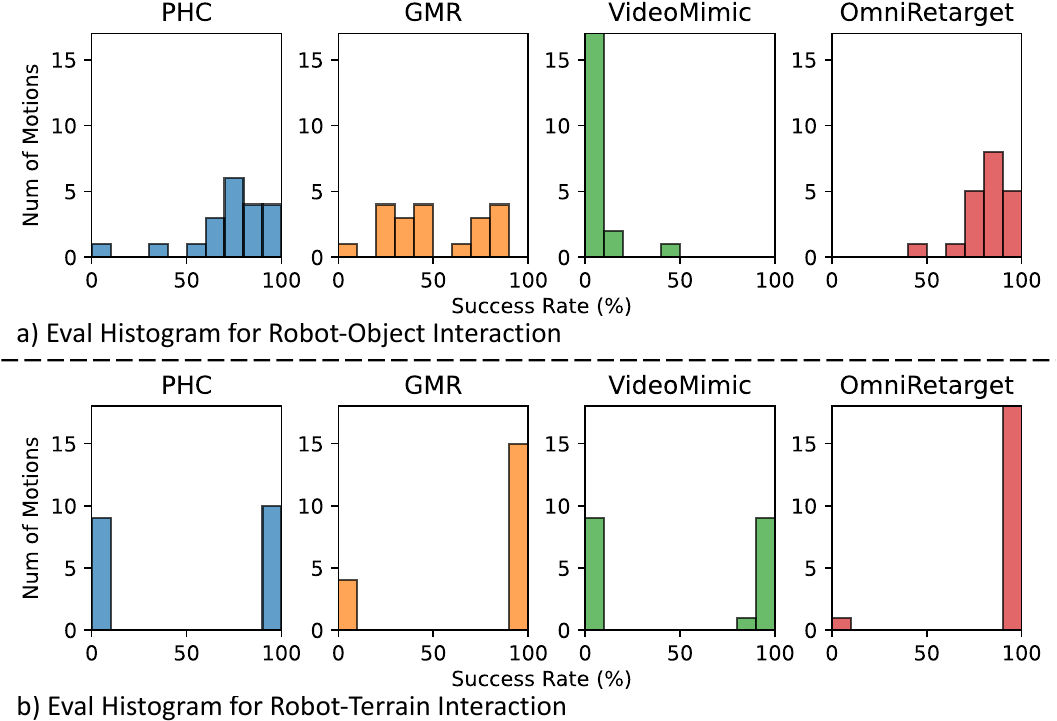}
	\caption{Histograms from the downstream RL evaluation showing the failure patterns for the baselines in different tasks. }
	\label{fig:histogram}
    \vspace*{-0.4cm}
\end{figure}

Shown in Fig.~\ref{fig:histogram}, we present histograms from the downstream RL evaluation (Sec.~\ref{sec:downstream-rl}) to illustrate failure patterns and variance across OmniRetarget and baselines. These histograms break down failure rates by each motion for two tasks: robot–object interaction and robot–terrain interaction, highlighting not only overall averages but also how failures distribute across different motions. We do not include augmented motions as baselines do not support augmentation. 

In robot–object interaction, the motions are modest while object properties are heavily randomized. Since the motions are not aggressive, most policies can adapt even to low-quality references and achieve at least one success, except VideoMimic, which fails systematically due to poor interaction preservation. This task therefore measures robustness rather than accuracy. We see that GMR shows broader failure spread with lower success rates, likely due to penetration issues that reduce robustness under placement changes. PHC shows improved robustness, while OmniRetarget achieves the most robust performance, with results concentrated in the high-success region.

In contrast, climbing terrains requires much more agile and challenging motions and thus, demands precise reference motions: if the quality is low, the agent fails outright with no successes. Here, PHC and VideoMimic perform the worst, with nearly half the motions failing entirely. GMR delivers somewhat better references but still fails on four motions, while OmniRetarget fails on only one. These results show that OmniRetarget not only provides superior robustness under variation but also higher reference accuracy.

For the one remaining failure, we believe that it is limited by the simple RL formulation we use. For future work, an interesting direction could be to extend the current RL formulation with curriculum learning to support these extremely difficult motions. 